\documentclass[conference]{IEEEtran}
\usepackage{xcolor,array}
\usepackage[colorlinks,urlcolor=orange,linkcolor=orange,citecolor=black]{hyperref}

\usepackage{amsmath,amssymb}
\usepackage{graphicx}
\usepackage{subcaption}
\usepackage{comment}
\usepackage[nolist]{acronym}

\usepackage[ruled]{algorithm2e}

\newcommand{\idea}[1]{#1}
\newcommand{\D}{\mathcal{D}}

\begin{document}
\bstctlcite{IEEEexample:BSTcontrol}

\title{B-PL-PINN: Stabilizing PINN Training with Bayesian Pseudo Labeling} 

\author{\IEEEauthorblockN{Kevin Innerebner\IEEEauthorrefmark{1}, Franz M. Rohrhofer\IEEEauthorrefmark{2}, and Bernhard C. Geiger\IEEEauthorrefmark{3}\IEEEauthorrefmark{2}}
\IEEEauthorblockA{\IEEEauthorrefmark{1}Institute of Human-Centred Computing, Graz University of Technology, Graz, Austria
}
\IEEEauthorblockA{\IEEEauthorrefmark{2}Know Center Research GmbH, Graz, Austria
}
\IEEEauthorblockA{\IEEEauthorrefmark{3}Signal Processing and Speech Communication Laboratory, Graz University of Technology, Graz, Austria\\
Email: innerebner@tugraz.at, frohrhofer@acm.org, geiger@ieee.org}
}

\maketitle

\begin{abstract}
Training physics-informed neural networks (PINNs) for forward problems often suffers from severe convergence issues, hindering the propagation of information from regions where the desired solution is well-defined. Haitsiukevich and Ilin (2023) proposed an ensemble approach that extends the active training domain of each PINN based on i) ensemble consensus and ii) vicinity to \mbox{(pseudo-)labeled} points, thus ensuring that the information from the initial condition successfully propagates to the interior of the computational domain. 

In this work, we suggest replacing the ensemble by a Bayesian PINN, and consensus by an evaluation of the PINN's posterior variance. Our experiments show that this mathematically principled approach outperforms the ensemble on a set of benchmark problems and is competitive with PINN ensembles trained with combinations of Adam and LBFGS.
\end{abstract}

\begin{IEEEkeywords}
Physics-informed neural networks; Bayesian pseudo labeling; partial differential equations; 
\end{IEEEkeywords}

\begin{acronym}
\acro{BPINN}{Bayesian Physics-Informed Neural Network}
\acro{PINN}{Physics-Informed Neural Network}
\acro{IC}{initial condition}
\acro{BC}{boundary condition}
\acro{BNN}{Bayesian Neural Network}
\acro{PDE}{partial differential equation}
\acro{NN}{Neural Network}
\acro{MCMC}{Markov chain Monte Carlo}
\acro{HMC}{Hamiltonian Monte Carlo}
\acro{NUTS}{No U-Turn Sampler}
\acro{PDF}{probability density function}
\acro{LHS}{latin hyper-cube sampling}
\acro{PSO}{particle swarm optimization}
\acro{MLE}{maximum likelihood estimation}
\end{acronym}

\section{Introduction}
\acp{PINN}~\cite{raissi2019physics} have become one of the cornerstones and main representatives of physics-informed machine learning due to their ability to incorporate differential equations during training. They were shown to be capable of solving both forward problems—simulating a system of differential equations for a given computational domain—and inverse problems, which involve determining problem parameters or unobserved variables from measurements.

A substantial portion of the scientific literature on \acp{PINN} is concerned with forward problems, or with certain variations thereof. In this case, training is mostly guided by the system of (ordinary or partial) differential equations as well as by the provision of \acp{IC}, \acp{BC}, and potentially additional data in the interior of the computational domain. The task here is to find a solution to the system of differential equations that satisfies \acp{IC} and \acp{BC} and agrees with the provided data. This forward problem has been shown to be particularly difficult to solve using \acp{PINN}, as \ac{PINN} training in this setting is characterized by a peculiar loss landscape, cf.~\cite{rohrhofer2023fixed}. Indeed, it has been argued that learning with such a loss landscape often leads to failures in the propagation of information from \acp{IC}, \acp{BC}, and measurement data to the interior of the computational domain~\cite{daw2023mitigating}. 

As a consequence, recent years have seen a surge in literature on stabilizing \ac{PINN} training (see also Section~\ref{sec:related:improvements}). Among these approaches is an ensemble method proposed in~\cite{haitsiukevich2023improved}, in which training an ensemble of \acp{PINN} is restricted to regions in the computational domain that are close to known, well-defined solution values (e.g., \acp{IC} or otherwise labeled points). If the ensemble reaches consensus for a training point in this region, then this point is labeled and used as anchor for subsequent iterations. As such, the proposed approach ensures reliable propagation of information from the boundary to the interior of the computational domain.

\begin{figure}
    \centering
    \includegraphics[width=\linewidth]{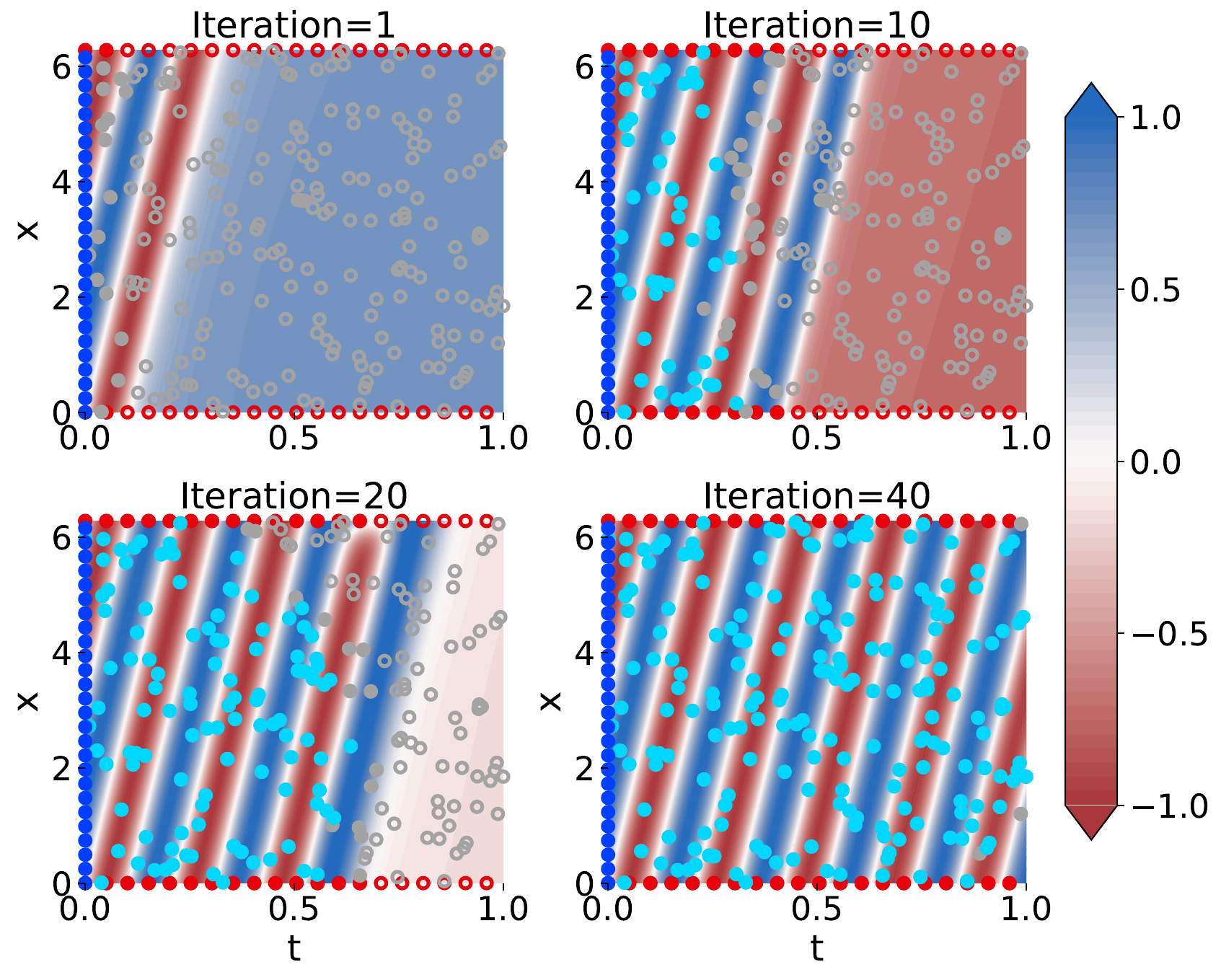}
    \caption{The solution provided by our B-PL-PINN for the convection system with $\beta=30$. Data instances for the initial condition and pseudo labels are shown in dark and light blue, collocation points in the interior and the boundary are shown in gray and red. Points are filled if used for updating the posterior at the current iteration. See Section~\ref{sec:method} for details.}
    \label{fig:training-region-expansion}
\end{figure}

In this work, we replace the ensemble from~\cite{haitsiukevich2023improved} by a Bayesian \ac{PINN}, effectively sampling from the posterior over solution functions and replacing ensemble consensus with a threshold on the posterior variance (Section~\ref{sec:method}). Thus, instead of training multiple \acp{PINN}, our approach relies on sampling a Bayesian \ac{PINN}. Our experiments with a set of benchmark problems (Section~\ref{sec:experiments}) show that our approach in many cases outperforms the ensemble when the latter is trained with Adam, and is competitive even when a combination of Adam and LBFGS is used for ensemble training (Section~\ref{sec:results}).  
Although our method relies on computationally expensive \ac{MCMC} sampling~\cite{neal2011mcmc,hoffman2014no}, our analysis shows that its runtime is not more than $9\times$ the runtime of the ensemble approach by Haitsiukevich and Ilin~\cite{haitsiukevich2023improved}.


Our proposed method\footnote{Code can be found at \url{https://github.com/kev-inn/B-PL-PINN}.} is a mathematically principled alternative to ensemble methods in \acp{PINN} and can be combined with a variety of complementary approaches to training stabilization, such as loss and collocation point weighting, domain decomposition, or curriculum learning (Section~\ref{sec:related:improvements}). As such, our work extends the toolbox of available training stabilization approaches and contributes towards successfully solving differential equations with \acp{PINN}.


\section{Background and Related Work}
\label{sec:related}
In this work we are concerned with using \acp{PINN} for solving systems of \acp{PDE}, that are given by:
\begin{equation}
\label{eq:system-pde}
\frac{\partial u}{\partial t} = \mathbf{\mathcal{N}}[u;\lambda], \quad x\in\Omega, t\in [0, T],
\end{equation}
where $u\equiv u(x,t)$ denotes the solution function, and $\mathcal{N}$ is a non-linear differential operator acting on $u$ and parameterized by $\lambda \in \mathbb{R}^P$. Furthermore, $\Omega\subseteq\mathbb{R}^D$ is the spatial domain, and $T$ the simulation time, defining the computational domain as $\Omega \times [0,T]$. 

In solving forward problems involving \acp{PDE}, \acp{IC} and \acp{BC} must be specified to clearly define the solution at the boundaries of the computational domain and to ensure the uniqueness of the solution. The \acp{IC} define the solution $u$ at the beginning of the simulation, expressed as $u(x, 0) = u_{ic}(x)$ for all $x \in \Omega$. The \acp{BC} employ a boundary operator $\mathbf{\mathcal{B}}$ (Dirichlet, Neumann, periodic) to enforce properties of the solution $u$ at the boundary $\partial\Omega$ of the spatial domain $\Omega$, i.e.,
\begin{equation}\label{eq:BC}
\mathbf{\mathcal{B}}[u]=0, \quad x\in \partial\Omega.
\end{equation}

\subsection{Physics-Informed Neural Networks}
\label{sec:related:PINN}
The idea of \acp{PINN}~\cite{raissi2019physics} is to use a \ac{NN} to approximate the true solution $u(x,t)$ to~\eqref{eq:system-pde} by a neural function $u_{\theta}(x,t)$ with (network) parameters $\theta$. In order to infer the solution function as specified by the \acp{IC} and \acp{BC}, \acp{PINN} are trained in a semi-supervised manner: labeled data $\D_l=\{(x_i,t_i,u_i)\}$, encoding measurements or simulation results, \acp{IC}, and/or fixed \acp{BC}, contribute to the supervised (data) loss component, while unlabeled \emph{collocation points} $\D_u=\{(x_i,t_i)\}$, which encode the governing \ac{PDE} and periodic \acp{BC}, are part of the unsupervised (physics) loss. The supervised (data) loss is given by the mean squared error between the labels and the results of the candidate solution $u_\theta$,
\begin{subequations}
    \begin{equation}\label{eq:dataloss}
        \mathcal{L}_{\theta}^{l}(\D_{l})=\frac{1}{|\D_l|} \sum_{i=1}^{|\D_l|}\lvert u_{\theta}(x_i, t_i) - u_i\rvert^2.
    \end{equation}
For example, a given initial condition $u_{ic}(x)$ yields data points $\{(x_i,0,u_{ic}(x_i))\}=\D_{ic}\subseteq \D_l$, while measurement data could provide a fully labeled dataset $\D_{data}$ from the entire computational domain.

The unsupervised (physics) loss evaluates a functional of the candidate solution $u_\theta$ as follows:,
    \begin{equation}\label{eq:physicsloss}
        \mathcal{L}_{\theta}^{u}(\D_{u})=\frac{1}{|\D_u|} \sum_{i=1}^{|\D_u|}\lvert \mathcal{G}[u_{\theta}](x_i, t_i)\rvert^2,
    \end{equation}
where the collocation points $\D_{pde}\subseteq \D_u$ are sampled from the interior of the computational domain.
This functional is evaluated using automatic differentiation~\cite{griewank2008evaluating} on the candidate function $u_\theta$, based on the definition of the physics residuals $f_{\theta}(x,t)$ (cf.~\eqref{eq:system-pde}):
\begin{equation}\label{eq:residual}
        \mathcal{G}[u_{\theta}](x,t) \triangleq f_{\theta}(x,t) = \frac{\partial u_{\theta}}{\partial t}(x,t) - \mathbf{\mathcal{N}}[u_{\theta}; \lambda](x,t).
\end{equation}
At the boundary of the computational domain, represented by the dataset $\D_{bc} \subseteq \D_u$, the unsupervised loss~\eqref{eq:physicsloss} is applied using the boundary residuals $b_\theta(x,t)$, defined as:
\begin{equation}\label{eq:BCloss}
    \mathcal{G}[u_{\theta}](x,t) \triangleq b_\theta(x,t) = \mathcal{B}[u_\theta](x,t).
\end{equation}
\end{subequations}

Training of \acp{PINN} thus involves minimizing a weighted combination of supervised and unsupervised losses, expressed as:
\begin{equation}\label{eq:PINNLoss_MO}
    \min_\theta \omega_l \mathcal{L}_{\theta}^{l}(\D_{l}) + \omega_u\mathcal{L}_{\theta}^{u}(\D_{u}),
\end{equation}
where $\omega_l$ and $\omega_u$ are the weighting factors for the respective loss components.
  
In many cases, particularly in forward problems, the losses~\eqref{eq:dataloss} and~\eqref{eq:physicsloss} are further divided into contributions from \acp{IC}, \acp{BC}, and the physics residual. This leads to the following formulation, as adopted in our work:
\begin{multline}\label{eq:PINNLoss}
    \mathcal{L}_{\theta}(\D_{ic}, \D_{bc}, \D_{pde}) \\=
    \omega_{ic}\mathcal{L}_{\theta}^{ic}(\D_{ic}) +
    \omega_{bc}\mathcal{L}_{\theta}^{bc}(\D_{bc}) +
    \omega_{pde}\mathcal{L}_{\theta}^{pde}(\D_{pde}).
\end{multline}

\subsection{Training Failures and Improvements of \ac{PINN} Training}\label{sec:related:improvements}
Since the initial employment of PINNs by~\cite{raissi2019physics}, numerous instances of training failures have been reported, prompting the development of various mitigation strategies discussed in the literature. One primary challenge stems from the multi-objective nature of PINN training (cf.~\eqref{eq:PINNLoss_MO}), which requires careful selection of loss weights $\omega$ to balance gradients and avoid training failures. Various schemes have been proposed to address this, including adaptive weighting strategies~\cite{wang2020understanding, xiang2022self}, manually chosen loss weights~\cite{rohrhofer2021pareto}, or hard constraints~\cite{lu2021physics,dong2021method}.
Another issue arises from the common initialization of PINNs and \acp{NN} in general, which can bias the model at the start of training toward trivial solutions of the physics loss function~\cite{leiteritz2021avoid} or fixed points in dynamical systems~\cite{rohrhofer2023fixed}. 
Architectural modifications, such as incorporating sinusoidal activation functions~\cite{LearningSinusoidal}, and various optimization reformulations have been proposed to mitigate these training challenges.
The latter include adaptive approaches like sequence-to-sequence/curriculum learning~\cite{krishnapriyan2021characterizing} and causality-aware weighting of collocation points~\cite{daw2023mitigating,wang2022respecting}, which specifically target issues related to what is known as the propagation failure. 
This failure is particularly critical in forward problem-solving, as it disrupts the propagation of information from \acp{IC} to the interior of the computational domain.
Domain decomposition approaches, such as those proposed in~\cite{jagtap2020extended} and~\cite{mattey2022novel}, are also widely used to mitigate propagation failures by training separate \acp{PINN} on smaller subsets of the temporal domain.


Most relevant to this work, however, are ensembles of \acp{PINN}—collections of individual \ac{PINN} instances used jointly to solve a given problem.
On the one hand, using such an ensemble helps mitigate issues in gradient-based optimization, e.g., by using \ac{PSO}~\cite{davi2022pso} where each \ac{PINN} instance represents a particle. While \ac{PSO} also incorporates gradient descent, it models swarm behavior, enabling ensemble members to share information and explore the optimization space more effectively.
On the other hand, \cite{haitsiukevich2023improved} used ensembles to directly tackle propagation failures during training. 
Specifically, the authors propose converting unlabeled collocation points to (pseudo-)labeled points when the ensemble shows agreement, and evaluating the physics loss~\eqref{eq:physicsloss} only for collocation points near labeled points.
By doing so, the algorithm adaptively expands the training domain in directions of consensus, starting from well-defined solution values, thereby ensuring the adequate propagation of \ac{IC}/\ac{BC} information to the interior of the computational domain.

\subsection{Contribution of This Work}
Our work builds on the ensemble approach of~\cite{haitsiukevich2023improved}, but replaces the ensemble with a \ac{BPINN}. 
\acp{BPINN} were introduced in~\cite{yang2021b} and have been shown to be effective in solving both forward and inverse problems when combined with \ac{MCMC} methods, such as \ac{HMC}.
Specifically, \acp{BPINN} enable uncertainty quantification, providing information about the model's confidence in the candidate solution $u_\theta$ at given points $(x_i,t_i)$ in the computational domain. 
As such, the \ac{BPINN} can be used to estimate consent in the sense of~\cite{haitsiukevich2023improved} by evaluating the variance of the posterior (see Section~\ref{sec:method} below). While there is often an implicit connection between ensemble methods such as~\cite{haitsiukevich2023improved} and Bayesian inference (e.g., the Bayes optimal classifier, Bayesian model averaging, and Bayesian model combining are all ensemble methods), our approach is explicitly Bayesian and mathematically well-founded.

Both our \ac{BPINN} approach and the ensemble approach of~\cite{haitsiukevich2023improved} are connected to approaches in~Section~\ref{sec:related:improvements}. The training domain is not extended based on residuals but based on epistemic uncertainty or consensus of the ensemble; collocation points are weighted not with real-valued weights, but with either 0 or 1, indicating whether they are active in the unlabeled loss~\eqref{eq:physicsloss}; collocation points are not explicitly activated based on their temporal position, but, as in causal approaches such as~\cite{wang2022respecting}, collocation points close to the \ac{IC} tend to be activated earlier.
Notably, since the training domain is based on epistemic uncertainty, regions of the spatial domain that are easier to learn may exhibit a higher activation rate, advancing more quickly to higher $t$ values compared to regions that are harder to learn.

\section{B-PL-PINNs: Bayesian Pseudo Labeling for Stable Training}
\label{sec:method}

The rationale behind B-PL-PINNs is that training \acp{PINN} for forward problems often fails because the \ac{IC} and/or \acp{BC} do not manage to propagate sufficiently to the interior of the computational domain, cf.~\cite{daw2023mitigating}. The ensemble approach of~\cite{haitsiukevich2023improved} thus incrementally expands the set of collocation points from known (labeled) points to those collocation points for which the ensemble $\{u_{\theta_i}:\forall i = 1, \ldots, N_{ensemble}\}$ of \ac{PINN} candidate solutions agrees. In our approach we replace the ensemble by a Bayesian \ac{PINN} (BPINN), and consider the agreement of an ensemble drawn from the posterior. In this section, we will describe our B-PL-PINN approach in detail. Pseudo code explaining the entire approach is presented in Algorithm~\ref{alg:main}.

\subsection{Mathematical Formulation}
\acp{BNN} explicitly specify the distribution of models we want to sample. Similar to an ensemble, it provides a measure of uncertainty, which is low in the close vicinity of (learned) training data, and high for regions not covered by the training data. In addition to the datasets $\D_{ic}$, $\D_{bc}$ and $\D_{pde}$ used for \ac{PINN} training, a dataset $\D_{pl}$ is therefore created during training to store pseudo-labels, expanding as the training progresses. The pseudo-labels are used to determine regions where the prediction is assumed to be accurate and the training domain can be further extended (see Section~\ref{sec:method_expanding}).

Our \ac{BPINN} employs Bayes' theorem to obtain a posterior distribution $P(\theta|\D)$ on the model parameters $\theta$ from a prior $P(\theta)$ and a likelihood $P(\D|\theta)$ given some data $\D$ via $P(\theta|\D)\propto P(\theta)P(\D|\theta)$. Here, the data $\D$ is given by both labeled data (such as from \acp{IC}, \acp{BC} or pseudo labels) and unlabeled data (corresponding to constraints imposing periodic \acp{BC} or \acp{PDE}).
In this work, the prior is considered Gaussian, i.e., $P(\theta) = \mathcal{N}(\theta; \mu_{p},\,\sigma_{p}^{2})$, where $\mu_{p}=0$ and $\sigma_{p}^2$ is a hyperparameter that can be adjusted. This is a common prior for \acp{BNN} that can be seen as a regularization on the network, similar to a $L_2$ regularization for gradient descent methods~\cite{silvestro2020prior}. 
For the likelihood, we also selected a Gaussian distribution both for the labeled and unlabeled datasets. Specifically, for $\D_l=\{(x_i,t_i,u_i)\}$, we set
\begin{subequations}\label{eq:likelihoods}
    \begin{equation}\label{eq:supervisedPosterior}
        P(\D_{l}\mid\theta) = \prod_{i=1}^{|\D_l|} \frac{1}{\sqrt{2\pi \sigma_{l}^{2}}} \exp{\left(- \frac{|u_{\theta}(x_i, t_i) - u_i|^2}{2\sigma_{l}^{2}}\right)}
    \end{equation}
while for unlabeled data points $\D_u=\{(x_i,t_i)\}$ at which the candidate solution must satisfy the constraints~\eqref{eq:residual} and/or \eqref{eq:BCloss}, we set
    \begin{equation}
        P(\D_{u}\mid\theta) = \prod_{i=1}^{|\D_u|} \frac{1}{\sqrt{2\pi \sigma_{u}^{2}}} \exp{\left(- \frac{|\mathcal{G}[u_\theta](x_i,t_i)|^2}{2\sigma_{u}^{2}}\right)}.
    \end{equation}
In our work, labeled data $\D_l$ consists of \ac{IC} points $\mathcal{D}_{ic}$ and pseudo-label points $\mathcal{D}_{pl}$, i.e., $\D_l=\D_{ic} \cup \D_{pl}$, with individual likelihood variances $\sigma_{ic}$ and $\sigma_{pl}$ for each dataset. 
On the other hand, unlabeled data $\D_u$ consists of periodic boundary condition points $\mathcal{D}_{bc}$ and collocation points $\mathcal{D}_{pde}$, i.e., $\D_u=\D_{bc} \cup \D_{pde}$, with again individual likelihood variances $\sigma_{bc}$ and $\sigma_{pde}$ assigned to each.
These likelihood variances, like the prior variance, are hyperparameters that must be selected before training.

As a consequence, the posterior of the model parameters given the full data $\D=\D_l \cup \D_u$ is proportional to 
\begin{equation}
\label{eq:posterior}
    P(\theta\mid \D) \approx P(\D_{l}\mid \theta) P(\D_{u}\mid \theta)  P(\theta).
\end{equation}
\end{subequations}
Notably, both the labeled $\D_l$ and unlabeled datasets $\D_u$ evolve throughout training, as discussed in Section~\ref{sec:method_expanding}.

\subsection{Sampling from the Posterior}
We use \ac{MCMC} to sample from the posterior $P(\theta\mid\mathcal{D})$. 
Since \ac{MCMC} does not require the posterior to be normalized, we can directly use the proportionality in~\eqref{eq:posterior} for our purpose. Specifically, we consider both \ac{HMC}~\cite{neal2011mcmc} with leapfrog integration and \ac{NUTS}~\cite{hoffman2014no}. 
Both options require setting the number of chains $N_{chains}$ and the number $N$ of samples to draw from the posterior; 
since \ac{MCMC} may be initialized at a low-probability region of the posterior, a certain number of $N_{burnin}$ samples will be drawn but discarded. Using the same approach as in~\cite{hoffman2014no}, the stepsize (e.g., of the leapfrog integrator) can be automatically learned by providing a target acceptance rate $\delta_{acc}$ that defines the trade-off between exploitation (staying close to the mode of the posterior) and exploration.


\subsection{Expanding the Training Region and Pseudo Labeling}
\label{sec:method_expanding}
Central to our algorithm is the adaptive expansion of the training domain. Intuitively, this is done by fixing the solution $u_{\theta_i}$ for points $(x_i,t_i)$ where the \ac{BPINN} is confident, adding these points to labeled data $\D_{pl}$ (pseudo labels). 
In addition to this expansion dynamic, only collocation points in $\D_{pde}$ that are near labeled data are considered during training.

For the expansion of the training domain and for the creation of pseudo labels, we iterate through the following five steps:
\begin{enumerate}
    \item Draw $N$ samples of model parameters $\{\theta_l\}_{l=1}^N$ from the posterior~\eqref{eq:posterior} with \ac{MCMC} and store them as a model ensemble
    \item Evaluate the ensemble's variance and transfer (unlabeled) points from $\D_{pde}$ to $\D_{pl}$ as new pseudo-labels if the ensemble reaches consensus and they are close to labeled data points (see below)
    \item Add new collocation points to $\D_{pde}$ within $\Delta_{pde}$ distance to the (new) pseudo labels
    \item Add training boundary points to $\D_{bc}$ within $\Delta_{pde}$ distance to the (new) pseudo labels
    \item Set \ac{MCMC} to the last model parameter sample $\theta_N$.
\end{enumerate}

Pseudo labels in step 2) are created in the following way: for an unlabeled collocation point $(x_i,t_i)\in\D_{pde}$, we search the closest labeled data point $(x_j,t_j,u_j)\in\D_l=\D_{ic}\cup\D_{pl}$. If this point satisfies 
\begin{subequations}
    \begin{align}
        \Vert (x_i,t_i)-(x_j,t_j)\Vert_2 &\le\Delta \label{eq:pl:dist}\\
        \left| u_j - \frac{1}{N}\sum_{\ell=1}^N u_{\theta_\ell}(x_j,t_j) \right| &\le \epsilon\\
        \frac{1}{N}\sum_{\ell=1}^N u_{\theta_\ell}^2(x_i,t_i) - \left(\frac{1}{N}\sum_{\ell=1}^N u_{\theta_\ell}(x_i,t_i)\right)^2 &\le \sigma_{consens}^2
    \end{align}
\end{subequations}
then a new pseudo label is created and added to $\D_{pl}$, i.e., $\D_{pl}= \D_{pl}\cup(x_i,t_i,u_i)$, where 
\begin{equation}
    u_i=\frac{1}{N}\sum_{\ell=1}^N u_{\theta_\ell}(x_i,t_i).
\end{equation}
In other words, we require that three conditions hold: i) The considered collocation point cannot be too far away from the closest labeled data point (hyperparameter $\Delta$). ii) The closest labeled data point must be reliable in the sense that its label agrees with the ensemble prediction (hyperparameter $\epsilon$). iii) The current collocation point must be such that the ensemble has achieved consensus in the sense that the variance of the ensemble is small (hyperparameter $\sigma_{consens}^2$).

Note that the distances in steps 3) and 4) and in~\eqref{eq:pl:dist} require the computation of a Euclidean distance between vectors describing the position in the computational domain. To make this meaningful, it is important to compute these distances from normalized coordinates, i.e., each dimension must be scaled to $[0,1]$.

\section{Experimental Setup}
\label{sec:experiments}

\subsection{Candidate Systems}
We follow the experimental setup of~\cite{haitsiukevich2023improved} and conduct experiments on four 1D systems of \acp{PDE} on a computational domain of $x\in[0,2\pi)$ and $t\in[0,1]$; we leave the evaluation of our approach for higher-dimensional \acp{PDE} for future work. Specifically, we consider a reaction system
\begin{subequations}
    \begin{equation}\label{eq:reaction}
        \frac{\partial u}{\partial t} = \rho u (1 - u),\quad u(x, 0) = \exp\left(-8\frac{(x - \pi)^2}{\pi^2}\right),
    \end{equation}
a diffusion system
    \begin{equation}\label{eq:diffusion}
        \frac{\partial u}{\partial t} =  \frac{1}{d^2} \frac{\partial^2 u}{\partial x^2},\quad 
        u(x, 0) = \sin(dx),
    \end{equation}
a reaction-diffusion system
    \begin{equation}\label{eq:reaction-diffusion}
        \frac{\partial u}{\partial t} = d \frac{\partial^2 u}{\partial x^2} + \rho u (1 - u),\quad 
        u(x, 0) = \exp\left(-8\frac{(x - \pi)^2}{\pi^2}\right),
    \end{equation}
and a convection system
    \begin{equation}\label{eq:convection}
        \frac{\partial u}{\partial t} = -\beta \frac{\partial u}{\partial x},\quad 
        u(x, 0) = \sin(x).
    \end{equation}
\end{subequations}
For all experiments we consider periodic Dirichlet \acp{BC}, i.e., $u(0, t) = u(2\pi, t)$; for the diffusion and the reaction-diffusion systems we additionally impose periodic Neumann \acp{BC}, i.e., $\frac{\partial u}{\partial x}(0, t) = \frac{\partial u}{\partial x}(2\pi, t)$. Each system is simulated with two different parameterizations ($\rho$, $d$, $\beta$).

\subsection{Data Creation}

Given an \ac{IC} dataset $\mathcal{D}_{ic}$ ($\vert\mathcal{D}_{ic}\vert=256$) and a training domain $\Omega$, we use \ac{LHS} to generate collocation points $\mathcal{D}_{pde}$ ($\vert\mathcal{D}_{pde}\vert=1\,000$) within $\Omega$. For the \ac{BC}, evenly spaced points are sampled on the boundary dimension and stored it in the boundary dataset $\mathcal{D}_{bc}$ ($\vert\mathcal{D}_{bc}\vert=100$). Additionally, we create an empty buffer dataset, $\D_{pl}$, which we will be used to store the pseudo-labels for collocation points where our prediction is likely close to correct. The pseudo-labels buffer will grow over time, but the size will stay in the range $0\le\vert\mathcal{D}_{pl}\vert\le\vert\mathcal{D}_{pde}\vert$.

\subsection{Baselines and Hyperparameter Settings}
\label{sec:experiments:hyperparameters}
We compare our proposed method with a vanilla \ac{PINN}, trained with the unmodified training procedure~\cite{raissi2019physics}, and our reimplementation of the ensemble method proposed in~\cite{haitsiukevich2023improved}, both with and without pseudo-labeling (see Section~\ref{sec:method}). 
In the latter case, pseudo-labels are not used to evaluate the supervised losses~\eqref{eq:dataloss} or~\eqref{eq:supervisedPosterior}, but only to act as ``anchors'' for extending the training domain of $\D_{pde}$. 
As neural architecture we choose a multi-layer perceptron with four hidden layers, $50$ neurons each, and $tanh$ activation function. 
The ensemble approach and our B-PL-PINN are trained for different numbers of iterations of Algorithm~\ref{alg:main}: For the reaction and reaction-diffusion system, training is conducted for 60 iterations; 
for the diffusion system 80 iterations are used for $d=5$, while 100 iterations are used for $d=10$; 
and for the convection system 100 iterations for $\beta=30$ and 150 iterations for $\beta=40$. 
We further train the vanilla \ac{PINN} and the ensemble method of~\cite{haitsiukevich2023improved} using a combination of Adam~\cite{kingma2014adam}  (learning rate $0.001$) and LBFGS~\cite{liu1989limited}. 
Specifically, in each iteration using the ensemble method, up to 20000 steps of LBFGS are taken after 5000 epochs of the Adam optimizer. 
For the vanilla \ac{PINN}, the same number of LBFGS steps are taken at the very end of the training using the Adam optimizer.
Furthermore, the vanilla \ac{PINN} uses uniform loss weights ($\omega_{ic}=\omega_{bc}=\omega_{pde}=1$), while the ensemble method uses $\omega_{ic}=1, \omega_{bc}=\frac{\vert \D_{bc}'\vert}{\vert \D_{bc}\vert}, \omega_{pde}=\frac{\vert \D_{pde}'\vert}{\vert \D_{pde}\vert}$.

To obtain an initial sample for the B-PL-PINN in the first iteration—assuming a solution to~\eqref{eq:PINNLoss} is also likely in the posterior formulation~\eqref{eq:posterior}—we first train a vanilla \ac{PINN} for 4000 Adam epochs and then initialize MCMC using the just obtained model parameters $\theta_{init}$ (cf. Algorithm~\ref{alg:main}). 
This allows us to reduce the number of burn-in samples for the first iteration, while for subsequent iterations the parameters from the previous iteration are re-used (see Section~\ref{sec:method_expanding}). 

As mentioned in Section~\ref{sec:method_expanding}, we perform min-max scaling to $[0,1]$ of the space and time coordinates to ensure that the hyperparameter settings $\Delta$ and $\Delta_{pde}$ do not depend on absolute scales. We use Optuna~\cite{akiba2019optuna} to obtain an initial set of hyperparameters for B-PL-PINN, such as the likelihood variances of~\eqref{eq:likelihoods} and the parameters of \ac{MCMC}. 
To this end, we run Optuna on the convection system~\eqref{eq:convection} for $\beta=30$ for approximately 100 trials. We employ the Tree-structured Parzen Estimator algorithm for hyperparameter sampling and prune trials with relative $L_2$ error above the median of previous trials at the current iteration.
Starting from the identified values (see Table~\ref{tab:hyperparameters}), we further manually adjust the hyperparameters due to following considerations: 
i) we use \ac{HMC} as \ac{NUTS} does not yield improvements over normal \ac{HMC}, but has worse runtime;
ii) the number of leapfrog steps $N_{leapfrog}$ is halved to ensure the runtime of B-PL-PINN remains comparable to competing approaches; 
iii) the number of burn-in samples $N_{burnin}$ is increased to ensure high-quality samples from the posterior;
iv) two chains are sampled, as they demonstrate better performance than a single chain and improve the sample quality;
v) increasing the number of chains allows us to reduce the number of samples $N$. 
With these adjustments, the runtime complexity of our approach was kept below $6\times$ for PL mode---below $9\times$ for No-PL mode---that of the ensemble approaches~\cite{haitsiukevich2023improved} used for comparison (Section~\ref{ssec:runtime}).

\begin{algorithm}[h]
\DontPrintSemicolon
\SetKwComment{Comment}{/* }{ */}
\SetKwFunction{FRelevantTrainingPoints}{RelevantTrainingPoints}
\SetKwProg{Fn}{function}{}{}

\caption{Bayesian PL-PINN Training}
\label{alg:main}
\KwData{$\D_{ic}, \D_{bc}, \D_{pde}$ /* data points */}
$\D_{pl} := \emptyset$\;
Randomly initialize model parameters $\theta_{init}$\;
\While{not converged}{
    $\D_{l} := \D_{ic} \cup \D_{pl}$\;
    $\D_{bc}' := \{(x,t)\in \D_{bc} : \exists (x',t',u')\in\D_l : \lVert (x,t) - (x',t') \rVert_2 < \Delta_{pde}\}$\;
    $\D_{pde}' := \{(x,t)\in \D_{pde} : \exists (x',t',u')\in\D_l : \lVert (x,t) - (x',t') \rVert_2 < \Delta_{pde}\}$\;
    \If{first iteration}{
        $\theta_{init} := $ Minimize loss
        $\mathcal{L}_{\theta}(\D_{ic}, \D_{bc}', \D_{pde}')$~\eqref{eq:PINNLoss} with Adam for 4000 epochs
    }
    $\theta_1, \ldots, \theta_{N} :=$ Sample $N$ model parameters using \ac{MCMC}, starting from $\theta_{init}$, using $\D_{l}, \D_{bc}', \D_{pde}'$ to compute the posterior $P(\theta_l\mid\D)$~\eqref{eq:posterior}\;
    \For{$(x,t) \in \D_{pde}'$}{
        $u_l := u_{\theta_l}(x,t),\quad \forall l \in 1, \ldots, N$\;
        $\sigma^2 = \mathsf{variance}(u_1,\dots,u_N)$\;
        $u := \mathsf{median}(u_1, \ldots, u_{N})$\;
        $(x',t',u') := \underset{(x',t',u') \in \D_l}{\arg\min} \lVert (x,t) - (x',t') \rVert_2$\;
        $u'' := \mathsf{mean}\left(u_{\theta_1}(x', t'), \ldots, u_{\theta_{N}}(x', t')\right)$\;
        \If{$\sigma^2 < \sigma_{consens}^2 \land \lVert (x,t) - (x',t') \rVert_2 < \Delta \land \lVert u' - u'' \rVert_2 < \epsilon$}{
            $\D_{pl} := \D_{pl} \cup \{(x,t,u)\}$
        }
    }
    $\theta_{init} := \theta_{N}$\;
}
\end{algorithm}

\begin{table}
\centering
\caption{Hyperparameter settings for pseudo labeling (top section), sampling (upper middle), model parameterization (lower middle), and initialization (bottom). If a range is given, automatic search was performed in this range. The table shows the best setting obtained from Optuna~\cite{akiba2019optuna} --- with the smallest relative $L_2$ error --- and the manually adapted values used for the final evaluation (see Section~\ref{sec:experiments:hyperparameters}). 
}
\label{tab:hyperparameters}
\begin{tabular}{c|c||c|c}
  \textbf{Hyperparameter} & \textbf{Range} & \textbf{\emph{Optuna}} & \textbf{Used Value}
  \\ \hline \hline $\Delta$ & - & - & 0.05
  \\ \hline $\Delta_{pde}$ & - & - & 0.1
  \\ \hline $\epsilon$ & - & - & $1\mathrm{e}{-3}$
   \\ \hline $\sigma_{consens}^2$ & $1\mathrm{e}{-6}$ to $1\mathrm{e}{-3}$ & $1.872\mathrm{e}{-4}$ & $2\mathrm{e}{-4}$
  \\ \hline \hline Sampler & \ac{HMC} or \ac{NUTS} & \ac{NUTS} & \ac{HMC}
   \\ \hline  $\delta_{acc}$ & $0.5$ to $0.99$ & $0.56$ & $0.6$
  \\ \hline $N$ & $10$ to $250$ & $185$ & 100
  \\ \hline $N_{burnin}$ & $0$ to $50$ & $46$ & 100
  \\ \hline $N_{leapfrog}$ & $2^3$ to $2^8$ & $2^8 = 256$ & $2^7 = 127$
  \\ \hline $N_{chains}$ & $1$ to $3$ & $1$ & $2$
  \\ \hline \hline $\sigma_{p}$ & $0.01$ to $10$ & $5.681$ & $5$
  \\ \hline $\sigma_{ic}$ & $0.001$ to $1$ & $1.016\mathrm{e}{-3}$ & $0.001$
  \\ \hline $\sigma_{pl}$ & $0.001$ to $1$ & $6.308\mathrm{e}{-3}$ & $0.005$
  \\ \hline $\sigma_{bc}$ & $0.001$ to $1$ & $1.041\mathrm{e}{-3}$ & $0.001$
  \\ \hline $\sigma_{pde}$ & $0.001$ to $1$ & $9.802\mathrm{e}{-3}$ & $0.01$
 \\ \hline \hline Initial epochs & - & - & 4000
 \\ \hline learning rate & - & - & 0.001
\end{tabular}
\end{table}

\subsection{Evaluation Metrics}
We evaluate the accuracy of all candidate solutions $u_\theta$ using the relative $L_2$ error $\|u_\theta-u\|_2/\|u\|_2$, evaluated at randomly sampled points from the computational domain. For the ensemble methods and the B-PL-PINN, the computed error is obtained by setting $u_\theta$ to the mean of the ensemble predictions.

\section{Results}
\label{sec:results}
\subsection{Performance on Benchmark Systems}
\begin{table*}
\centering
\caption{Relative $L_2$ error for different candidate systems and \ac{PINN} methods. Bold numbers indicate best results with and without using LBFGS, respectively. The asterisk for the results of our method for diffusion with $d=10$ indicates that the number of epochs for which the weights were trained using~\eqref{eq:PINNLoss} had to be increased to 40.000.}
\label{tab:results}
\begin{tabular}{l||cc|cc|cc|cc}
  & \multicolumn{2}{c|}{Reaction} & \multicolumn{2}{c|}{Diffusion} & \multicolumn{2}{c|}{Reaction-Diffusion ($\rho=5$)}& \multicolumn{2}{c}{Convection}\\
   \textbf{Method}  &  $\rho=5$ & $\rho=7$ & $d=5$ & $d=10$ & $d=2$ & $d=4$ & $\beta=30$ & $\beta=40$\\
   \hline
   PINN  & 7.85e-03 & 9.73e-01 & 1.87e-02 & 5.31e-02 & 6.94e-01 & 8.88e-01  & 1.59e-02 & 1.56e-02\\
   \hline
   Ensemble PL & 2.12e-02 & 5.92e-02 & 2.03e-02 & 3.74e-02 & 8.78e-03 & 8.32e-03  & 2.43e-02 & 5.87e-02\\
   Ensemble Ens & 1.04e-02 & 2.44e-02 & 1.77e-02 & 7.52e-02 & 7.52e-03 & \textbf{7.22e-03}  & \textbf{1.19e-02} & 4.01e-02\\
   Bayesian PL (ours) &  \textbf{7.36e-03} & \textbf{1.14e-02} & \textbf{1.61e-02} & \textbf{2.48e-02}$^*$ & 7.62e-03 & 7.26e-03 & 1.43e-02 & 3.39e-02 \\
   Bayesian No-PL (ours) & 1.97e-02 & 4.33e-02 & 1.58e-02 & 2.59e-02$^*$  & \textbf{7.18e-03} & 7.82e-03 & 1.21e-02 & \textbf{1.17e-02}\\
   \hline
   PINN (w/ LBFGS) & 7.85e-03 & 9.73e-01 & 1.55e-02 & 2.41e-02 & 6.96e-01 & 8.88e-01 & 1.49e-02 & 1.49e-02 \\
   Ensemble PL (w/ LBFGS) & \textbf{7.40e-03} & 2.38e-02 & \textbf{1.54e-02} & 2.41e-02 & 7.35e-03 & \textbf{7.02e-03} & \textbf{1.02e-02} & \textbf{1.26e-02}  \\
   Ensemble Ens (w/ LBFGS) & 9.32e-03 & \textbf{7.02e-03} & \textbf{1.54e-02} & \textbf{2.36e-02}  & \textbf{7.31e-03} & 7.09e-03 & 1.26e-02 & 1.66e-02
\end{tabular}
\end{table*}

Table~\ref{tab:results} presents the quantitative results of our experiments, while Fig.~\ref{fig:training-region-expansion} illustrates an exemplary run for the reaction system with $\rho=5$. 
As apparent from the table, our B-PL-PINN outperforms the ensemble method in six out of the eight considered systems, resulting in relative $L_2$ errors of less than 5\% in all cases. Since it is known that LBFGS and other second-order methods are useful for \ac{PINN} training especially when combined with first-order methods like Adam~\cite{rathore2024challenges}, we also investigated the performance of a standard \ac{PINN} and the ensemble approach with this optimization approach included. While the usage of LBFGS in most cases led to superior results, also here the quantitative performance is within a similar range. In addition, LBFGS increases the computational complexity of the ensemble approach by a factor of up to 2.

Visual inspection of the error patterns reveals that the errors by our B-PL-PINN are qualitatively similar to the errors of the ensemble approach by~\cite{haitsiukevich2023improved}, reflecting the overall similarity of the quantitative results. Minor differences can be seen in the convection system, where the ensemble method appears to slightly misestimate the wave speed, while the accuracy of the B-PL-PINN becomes worse for larger temporal coordinates. Further, due to the use of the median operation in Algorithm~\ref{alg:main}, our solutions are generally less smooth than those from the ensemble approach, which is reflected in the error patterns. Finally, for the diffusion experiment with $d=10$, 4000 epochs of Adam were not sufficient to accurately learn the \ac{IC}. As a consequence, subsequent iterations of the algorithm did not yield satisfactory results, as the B-PL-PINN did not manage to find an appropriate mode of the posterior distribution. Increasing the number of initial epochs to 40000 (indicated with asterisk in Table~\ref{tab:results}), leads to a successfully learned \ac{IC} and to competitive performance. We thus conclude that learning the \ac{IC} successfully is essential for achieving adequate performance.

\subsection{Ablation Analysis}

\begin{figure}
    \centering
    \includegraphics[width=\linewidth]{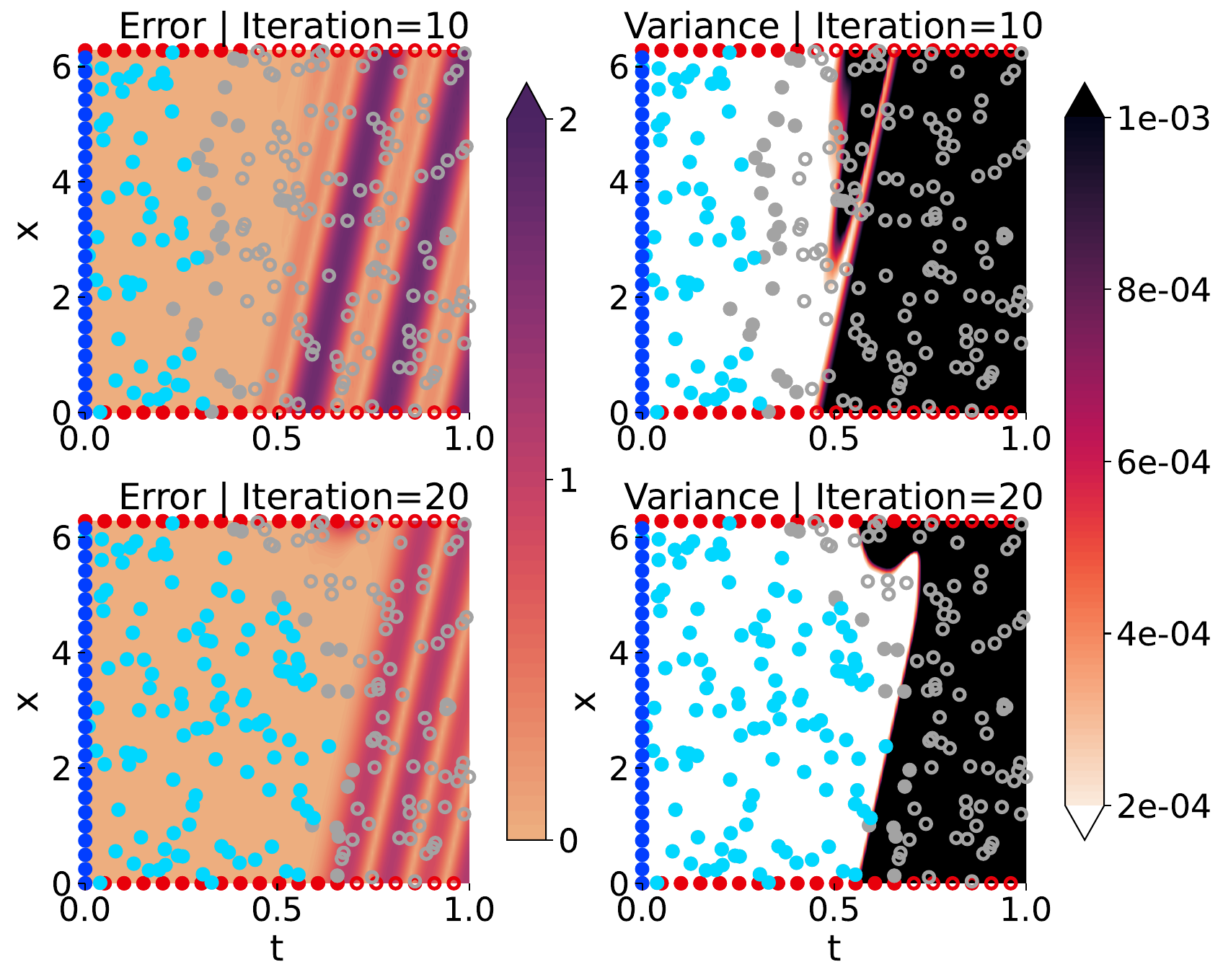}
    \caption{The absolute error (left) and posterior variance (right) of the B-PL-PINN predictions after the 10th iteration (top) and after the 20th iteration (bottom) on the convection system with $\beta=30$ (cf. Figure~\ref{fig:training-region-expansion}). Data instances for the initial condition and pseudo labels are shown in dark and light blue, collocation points in the interior and the boundary are shown in gray and red. Points are filled if used for updating the posterior at the current iteration.}
    \label{fig:uncertainty}
\end{figure}
\begin{figure}
    \centering
    \includegraphics[width=\linewidth]{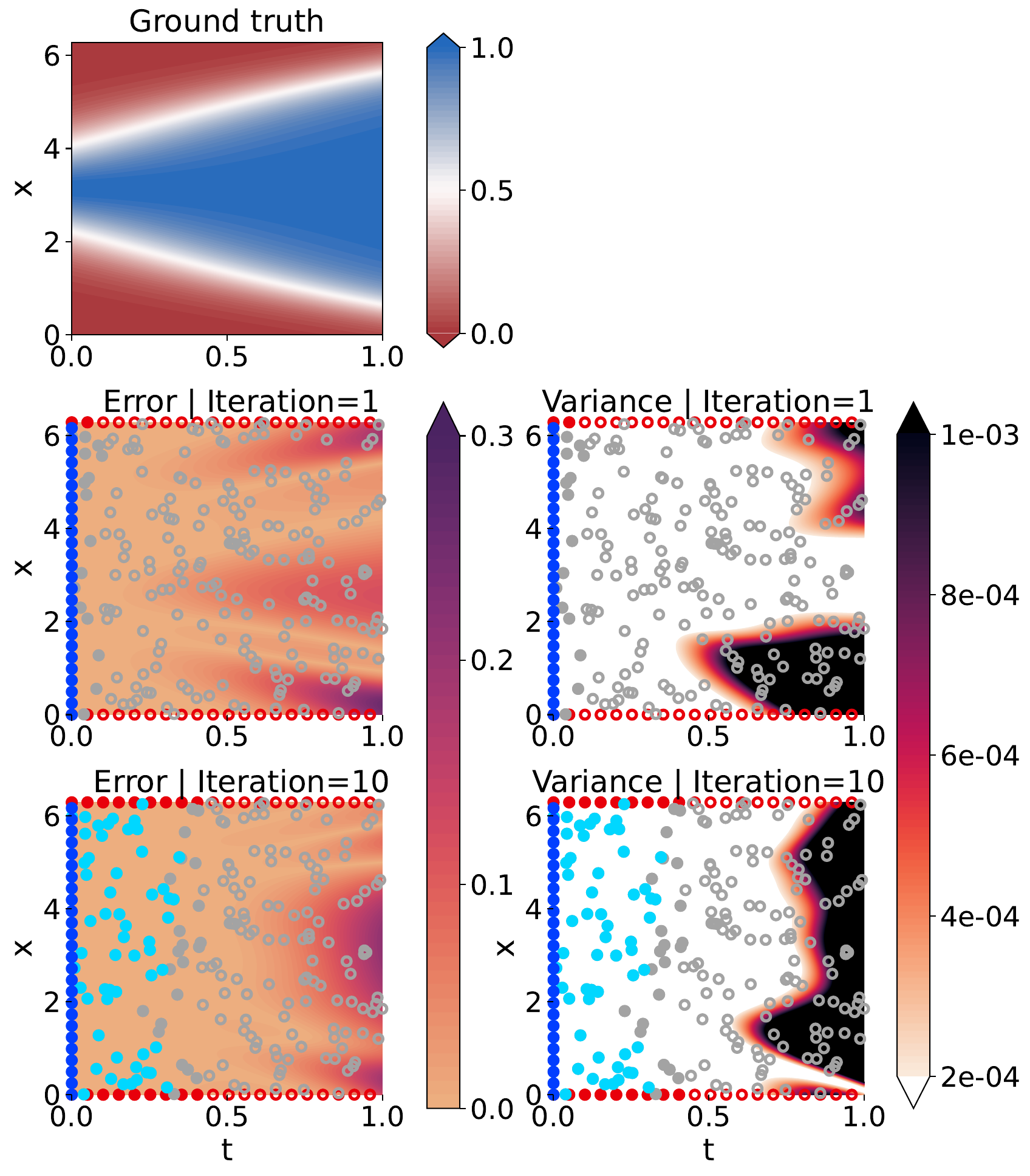}
    \caption{The ground truth (top left), absolute error (left) and posterior variance (right) of the B-PL-PINN predictions after the initial iteration (top) and after the 10th iteration (bottom) on the reaction system with $\rho=5$. Data instances for the initial condition and pseudo labels are shown in dark and light blue, collocation points in the interior and the boundary are shown in gray and red. Points are filled if used for updating the posterior at the current iteration.}
    \label{fig:uncertainty-reaction}
\end{figure}

The algorithm of our approach, as well as the one of~\cite{haitsiukevich2023improved}, depends on both consensus and closeness to labeled points. While the requirement of consensus is intuitive, we here show that it is not sufficient and that closeness to labeled data is an essential ingredient of our method. To this end, we visualize in Fig.~\ref{fig:uncertainty} the variance of the posterior~\eqref{eq:posterior} and the absolute error of the candidate solution.

For the convection system, the posterior variance is a good predictor for the error for the whole training domain (see Fig.~\ref{fig:uncertainty}). When the error is small, the variance is below our chosen threshold ($\sigma^2_{consens}=2\mathrm{e}{-4}$), when the error is high, the variance is above the threshold.
Therefore, \idea{though it has been shown that Bayesian Neural Networks can underestimate uncertainty far away from their training points~\cite{DBLP:journals/corr/abs-2010-02709},} in this case, 
the variance appears to be a sufficient condition for labeling.

In contrast to the convection system, the iterative propagation behaves differently on the reaction system.  
\idea{As it can be seen, after the initial iteration --- the \ac{BPINN} was trained on \ac{IC} for the \emph{data loss} and the \emph{physics loss} evaluated for collocation and boundary points within $\Delta_{pde}$ distance --- the posterior variance is below $\sigma^2_{consens}=2\mathrm{e}{-4}$ for most of the domain (see Fig.~\ref{fig:uncertainty-reaction}, middle right), especially in the middle of the spatial domain (approximately $x\in[2,4]$).
Although, the estimated uncertainty is low for these regions, there are still noticeable errors (see Fig.~\ref{fig:uncertainty-reaction}, middle left).
While after 10 iterations the errors align better with the estimated uncertainty, there are still areas with large errors for which the posterior variance is small (e.g., the region at $x\in[0,2], t\in[0.75,1]$ in Fig.~\ref{fig:uncertainty-reaction}, bottom right).} This indicates that relying on consensus alone is insufficient, more so for some systems than others.
As a consequence, the speed of the training domain expansion, can be limited both by the parameter $\Delta$, or $\sigma^2_{consens}$, depending on the system in question.


\subsection{Runtime Analysis}
\label{ssec:runtime}

We list the end-to-end runtimes of the methods, with default configurations, on the reaction system experiment with $\rho=5$ in Table~\ref{tab:runtime}. We did not implement notable optimizations and run experiments on an AMD Ryzen 5 3600X 6-Core Processor (3.80 GHz) CPU.
Our runtimes are worse, but in a similar range as the ensemble methods (below $6\times$ for PL and below $9\times$ for No-PL).
In addition to the algorithmic differences between the ensemble and Bayesian methods, the methods are implemented in different frameworks.
While our method is implemented in Tensorflow 2~\cite{tensorflow2015-whitepaper} with Tensorflow Probability~\cite{DBLP:journals/corr/abs-1711-10604}, the ensemble method uses PyTorch~\cite{DBLP:conf/nips/PaszkeGMLBCKLGA19}.

The number of iterations is fixed at $60$ for the ensemble methods and Bayesian methods. Both methods usually converge earlier for this experiment, with all, or most, collocation points activated and with a pseudo-label assigned. Because the number of operations depends on the amount of active collocation points and pseudo-labels, more time is spent on later iterations. Therefore, stopping once the training domain is explored, could substantially reduce runtime for both methods.

\begin{table}
\centering
\caption{Runtimes of methods on the reaction system with $\rho=5$.}
\label{tab:runtime}
\begin{tabular}{l||c|c}
  \textbf{Method} & Runtime & Factor (Runtime / Ensemble PL)\\
   \hline
   PINN   &  ~~7m54s  & 0.41 \\ \hline
   Ensemble PL  &  ~19m18s  & 1.00 \\
   Ensemble Ens  &  ~18m49s  & 0.97 \\
   Bayesian PL (ours)  &  109m36s  & 5.68 \\
   Bayesian No-PL (ours)  &  165m23  & 8.57  \\ \hline
   PINN (w/ LBFGS)  &  ~~8m11s  & 0.42 \\
   Ensemble PL (w/ LBFGS)  &  ~28m12s & 1.46 \\ 
   Ensemble Ens (w/ LBFGS)  &  ~26m39s & 1.38 \\
\end{tabular}
\end{table}

\section{Conclusion}
In this work, we propose a Bayesian approach for iteratively expanding the relevant computational domain for \ac{PINN} training. Based on a recently proposed ensemble approach, the performance of our B-PL-PINNs is competitive both in terms of accuracy and computational complexity.

Future work shall investigate ways to improve the efficiency of \ac{MCMC} by sampling only parameters from a subset of layers, as in~\cite{zeng2018relevance,brosse2020last}, and by combining our approach with methods for adaptive collocation point sampling.
Although previous research has shown advantages of \ac{MCMC} methods over variational inference~\cite{yang2021b} for Bayesian PINNs, future work may explore a variational inference implementation of our method to assess potential runtime improvements.

\section*{Acknowledgments}
This work was supported by the European Union’s HORIZON Research and Innovation Programme under grant agreement No 101120657, project ENFIELD (European Lighthouse to Manifest Trustworthy and Green AI).

\bibliographystyle{IEEEtran}
\bibliography{references}

\end{document}